\documentclass[conference]{IEEEtran}
\IEEEoverridecommandlockouts
\usepackage{cite}
\usepackage{amsmath,amssymb,amsfonts}
\usepackage{algorithmic}
\usepackage{graphicx}
\usepackage{textcomp}
\usepackage{makecell}
\usepackage{xcolor}
\def\BibTeX{{\rm B\kern-.05em{\sc i\kern-.025em b}\kern-.08em
    T\kern-.1667em\lower.7ex\hbox{E}\kern-.125emX}}
\begin{document}

\title{Syn-TurnTurk: A Synthetic Dataset for Turn-Taking Prediction in Turkish Dialogues\\

}

\author{\IEEEauthorblockN{Ahmet Tuğrul Bayrak}
\IEEEauthorblockA{\textit{Data Science and Innovation} \\
\textit{Ata Technology Platforms}\\
İstanbul, Turkey \\
tugrul.bayrak@atptech.com}
\and
\IEEEauthorblockN{Mustafa Sertaç Türkel}
\IEEEauthorblockA{\textit{Data Science and Innovation} \\
\textit{Ata Technology Platforms}\\
İstanbul, Turkey \\
sertac.turkel@atptech.com}
\and
\IEEEauthorblockN{Fatma Nur Korkmaz}
\IEEEauthorblockA{\textit{Data Science and Innovation} \\
\textit{Ata Technology Platforms}\\
İstanbul, Turkey \\
fatmanur.korkmaz@atptech.com}
}

\IEEEoverridecommandlockouts
\IEEEpubid{\makebox[\columnwidth]{© 2026 IEEE. Personal use of this material is permitted. \hfill} \hspace{\columnsep}\makebox[\columnwidth]{ }}

\maketitle

\begin{center}
\small
This is the author's version of a paper accepted for publication in IEEE ICASI 2026.
\end{center}

\begin{abstract}
Managing natural dialogue timing is a significant challenge for voice-based chatbots. Most current systems usually rely on simple silence detection, which often fails because human speech patterns involve irregular pauses. This causes bots to interrupt users, breaking the conversational flow. This problem is even more severe for languages like Turkish, which lack high-quality datasets for turn-taking prediction. This paper introduces Syn-TurnTurk, a synthetic Turkish dialogue dataset generated using various Qwen Large Language Models (LLMs) to mirror real-life verbal exchanges, including overlaps and strategic silences. We evaluated the dataset using several traditional and deep learning architectures. The results show that advanced models, particularly BI-LSTM and Ensemble (LR+RF) methods, achieve high accuracy (0.839) and AUC scores (0.910). These findings demonstrate that our synthetic dataset can have a positive affect for models understand linguistic cues, allowing for more natural human-machine interaction in Turkish.
\end{abstract}


\section{Introduction}
The growth of generative AI has made chatbots a common tool in many industries. While creating a basic system is now straightforward, largely due to methods like Retrieval-Augmented Generation (RAG), developing a bot that interacts like a human remains a significant challenge. This difficulty is especially clear in voice-based systems, where the timing of a conversation is essential for a natural experience.

Most voice-enabled chatbots function by monitoring for a specific duration of silence. Once the user stops speaking for a few seconds, the system assumes the turn has ended and begins its response. However, this method is often unreliable because speech patterns vary between individuals. People frequently pause in the middle of a sentence or between words. If a bot interprets these pauses as a finished turn, it will interrupt the user. This mistake disturbs the conversational flow and makes the interaction feel mechanical rather than human.

A fundamental requirement for realistic spoken dialogue systems is the ability to manage turn-taking with human-like speed and accuracy. Recent surveys highlight that a major challenge in this field is the lack of established benchmarks to track progress and compare models on a standard ground \cite{b11}. Understanding the underlying structure of speech is essential for modeling these interactions. The GRASS corpus \cite{b14} established a framework for this by defining turn-taking in terms of layers such as Inter-Pausal Units (IPU) and Potential Completion Points (PCOMP). These linguistic markers served as a core motivation for TurnGPT \cite{b7}, a study that proved Transformer-based models could use dialogue context not only to detect turn changes but also to anticipate them. Building on this, "projection" mechanisms was introduced\cite{b9} designed to predict future completion points, which successfully reduced system latency and made interactions more immediate.

As the field moves toward more complex systems, models must handle signals that occur at different temporal speeds. Roddy et al. \cite{b10} addressed this by proposing a Multiscale RNN architecture that processes linguistic and acoustic features at separate rates. More recently, the scope of turn-taking has expanded to include multimodal signals. For instance, the MM-F2F dataset \cite{b13} combines linguistic, acoustic, and visual data to improve the prediction of both turn-taking and backchannel actions. Visual information is particularly useful when audio data is compromised or unavailable. In such scenarios, Cano et al. \cite{b12} demonstrated that social robots can rely on Visual Voice Activity Detection (VVAD) to identify speech boundaries and manage conversational flow using only visual cues.

The difficulty of predicting turns is even greater for languages that are not as widely supported in global datasets. Turkish, in particular, suffers from a lack of high-quality, labeled conversation data needed to train accurate turn-taking models. Due to its unique sentence structure and suffix-based grammar, existing models trained on English often fail to capture the nuances of Turkish dialogue. To address this gap, this paper presents a synthetic Turkish dialogue dataset. The goal is to train models that can recognize turn boundaries by analyzing the linguistic structure of Turkish conversations. By learning from these computer-generated examples, the models can perceive when a user has completed their thought based on the specific words and grammatical markers used. This approach moves away from a reliance on simple silence and toward a more intelligent, language-specific understanding of conversational flow.

\section{Dataset Generation}
The construction of this dataset involved the utilization of five distinct Qwen models via API calls: qwen3-max-2026-01-23, qwen3.5-35b-a3b, qwen3.5-plus-2026-02-15, qwen3.5-397b-a17b, and qwen3.5-flash-2026-02-23. Each model was tasked with generating natural, two-person dialogues based on specific constraints to ensure the dataset's diversity.

To prevent semantic repetition and ensure a broad coverage of conversational contexts, a pool of 79 unique topics was established. For each API request, a topic was selected at random, serving as the thematic foundation for the interaction. Furthermore, the models were specifically instructed to incorporate human-centric speech characteristics, such as overlaps, strategic silences, and everyday interjections, to mirror real-life verbal exchanges as closely as possible. The variability of the generated outputs was controlled by adjusting the temperature parameter. Although different values were used to increase diversity, most generations were performed at a temperature of 0.7 to balance coherence and conversational spontaneity. The resulting raw data (Syn-TurnTurk) was subsequently formatted and hosted on Hugging Face\footnote{https://huggingface.co/datasets/tugrulbayrak/Syn-TurnTurk} for further analysis.

\section{Dataset Characteristics and Structural Analysis}
A structural analysis was conducted to quantify the dataset's characteristics and ensure its suitability for training predictive models. The final corpus consists of 1,625 dialogues, with a total of 12,560 individual speaker changes. This high volume of turn-taking suggests that the interactions are sufficiently dynamic, moving beyond simple prompt-response pairs into more complex, multi-turn exchanges. The dataset's diversity is reflected in both its thematic breadth and its parametric variety. By distributing the dialogues across 79 distinct topics, the risk of linguistic over-fitting was minimized. The distribution of dialogues across the five Qwen models is detailed in Table \ref{tab_model_dist}.

\begin{table}[htbp]
\caption{Distribution of Dialogues by Model}
\begin{center}
\begin{tabular}{|l|c|c|}
\hline
\multicolumn{1}{|c|}{\textbf{Model Name}} & \textbf{Count} & \textbf{Percentage (\%)} \\
\hline
qwen3-max-2026-01-23 & 675 & 41.5\% \\
\hline
qwen3.5-35b-a3b & 283 & 17.4\% \\
\hline
qwen3.5-flash-2026-02-23 & 270 & 16.6\% \\
\hline
qwen3.5-397b-a17b & 228 & 14.0\% \\
\hline
qwen3.5-plus-2026-02-15 & 169 & 10.4\% \\
\hline
\end{tabular}
\label{tab_model_dist}
\end{center}
\end{table}




To evaluate the timing of the conversations, Floor Transfer Offset (FTO) and general interaction flow metrics were measured to capture the temporal nuances of human-like exchanges. These values, summarised in Table \ref{tab_timing_metrics}, represent the core temporal characteristics and the dynamic nature of the generated dialogues. As illustrated in Fig. 1, the difference between the mean and median FTO suggests an asymmetric distribution, with a significant number of longer silence gaps alongside frequent rapid transitions, which are critical for modelling natural speech patterns accurately. Specifically, the dataset contains 5,305 documented instances of overlaps, totalling 2,213.50 seconds of concurrent speech, reflecting a high level of interactivity within the synthetic interactions. These metrics provide a robust baseline for the subsequent evaluation of 30 different combinations of models, specifically testing their ability to distinguish between intentional pauses and actual completed turns.

\begin{table}[htbp]
\caption{Summary of Timing and Interaction Metrics}
\begin{center}
\begin{tabular}{|l|c|}
\hline
\multicolumn{1}{|c|}{\textbf{Metric}} & \textbf{Value} \\
\hline
Mean FTO & 0.286s \\
\hline
Median FTO & 0.743s \\
\hline
Max Overlap (Negative FTO) & -2.500s \\
\hline
Max Silence Gap (Positive FTO) & 0.880s \\
\hline
Total Number of Overlaps & 5,305 \\
\hline
Avg. Overlaps per Dialogue & 3.26 \\
\hline
Avg. Silence per Dialogue & 3.58s \\
\hline
\end{tabular}
\label{tab_timing_metrics}
\end{center}
\end{table}

\begin{figure}[htbp]
\centerline{\includegraphics[width=\columnwidth]{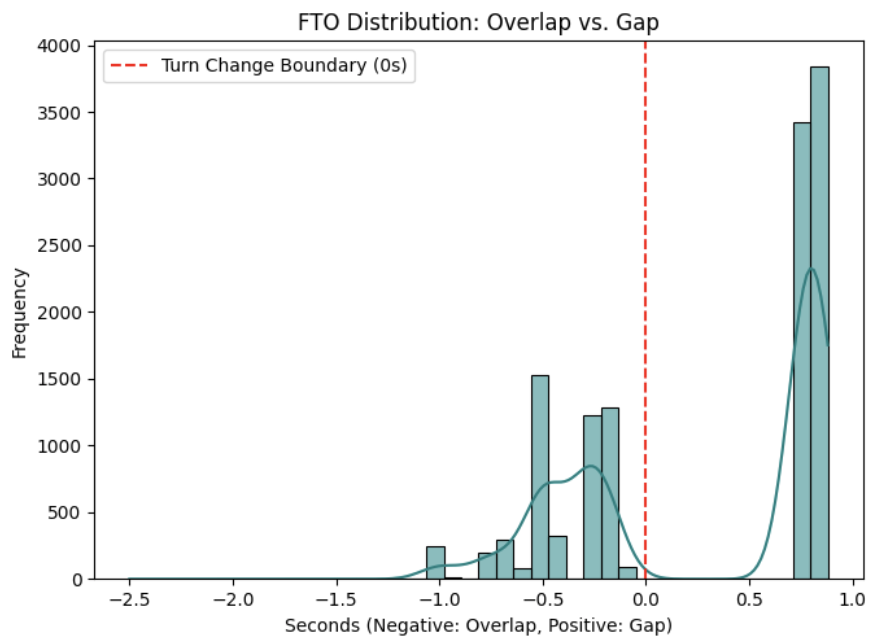}}
\caption{FTO distribution histogram}
\label{fto}
\end{figure}

\begin{table*}[t]
\footnotesize
\caption{Overall Performance Comparison: Absolute Best Model per Metric Highlighted}
\begin{center}
\setlength{\tabcolsep}{15pt} 
\begin{tabular}{|l|l|c|c|c|c|c|}
\hline
\textbf{Data Subset} & \textbf{Model} & \textbf{Precision} & \textbf{Recall} & \textbf{F1-Score} & \textbf{Accuracy} & \textbf{AUC} \\
\hline
Full Dataset & Logistic Regression & 0.697 & \textbf{0.845} & 0.764 & 0.816 & 0.898 \\
\cline{2-7}
 & Decision Tree & 0.597 & 0.600 & 0.598 & 0.721 & 0.687 \\
\cline{2-7}
 & Random Forest & 0.808 & 0.564 & 0.664 & 0.802 & 0.890 \\
\cline{2-7}
 & Ensemble (LR+RF) & 0.754 & 0.794 & \textbf{0.773} & 0.836 & 0.907 \\
\cline{2-7}
 & LSTM & 0.765 & 0.761 & 0.763 & 0.838 & 0.904 \\
\cline{2-7}
 & BI-LSTM & 0.767 & 0.778 & 0.772 & 0.838 & 0.905 \\
\hline
qwen3.5-397b-a17b & Logistic Regression & 0.661 & 0.813 & 0.729 & 0.794 & 0.882 \\
\cline{2-7}
 & Decision Tree & 0.571 & 0.596 & 0.584 & 0.709 & 0.634 \\
\cline{2-7}
 & Random Forest & 0.807 & 0.456 & 0.582 & 0.777 & 0.867 \\
\cline{2-7}
 & Ensemble (LR+RF) & 0.737 & 0.723 & 0.730 & 0.818 & 0.886 \\
\cline{2-7}
 & LSTM & 0.751 & 0.719 & 0.735 & 0.823 & 0.892 \\
\cline{2-7}
 & BI-LSTM & 0.752 & 0.689 & 0.719 & 0.816 & 0.889 \\
\hline
qwen3.5-plus-2026-02-15 & Logistic Regression & 0.677 & 0.821 & 0.742 & 0.806 & 0.887 \\
\cline{2-7}
 & Decision Tree & 0.552 & 0.597 & 0.574 & 0.698 & 0.631 \\
\cline{2-7}
 & Random Forest & \textbf{0.814} & 0.454 & 0.583 & 0.779 & 0.868 \\
\cline{2-7}
 & Ensemble (LR+RF) & 0.749 & 0.726 & 0.737 & 0.824 & 0.886 \\
\cline{2-7}
 & LSTM & 0.718 & 0.751 & 0.734 & 0.815 & 0.888 \\
\cline{2-7}
 & BI-LSTM & 0.742 & 0.672 & 0.705 & 0.809 & 0.887 \\
\hline
qwen3-max-2026-01-23 & Logistic Regression & 0.681 & 0.837 & 0.751 & 0.813 & 0.897 \\
\cline{2-7}
 & Decision Tree & 0.585 & 0.684 & 0.631 & 0.730 & 0.686 \\
\cline{2-7}
 & Random Forest & 0.811 & 0.544 & 0.651 & 0.804 & 0.893 \\
\cline{2-7}
 & Ensemble (LR+RF) & 0.747 & 0.785 & 0.766 & 0.838 & \textbf{0.910} \\
\cline{2-7}
 & LSTM & 0.786 & 0.710 & 0.746 & 0.837 & 0.906 \\
\cline{2-7}
 & BI-LSTM & 0.776 & 0.734 & 0.754 & \textbf{0.839} & 0.907 \\
\hline
qwen3.5-35b-a3b & Logistic Regression & 0.687 & 0.828 & 0.751 & 0.814 & 0.893 \\
\cline{2-7}
 & Decision Tree & 0.570 & 0.638 & 0.602 & 0.715 & 0.648 \\
\cline{2-7}
 & Random Forest & 0.805 & 0.470 & 0.594 & 0.782 & 0.874 \\
\cline{2-7}
 & Ensemble (LR+RF) & 0.733 & 0.733 & 0.733 & 0.819 & 0.893 \\
\cline{2-7}
 & LSTM & 0.761 & 0.743 & 0.752 & 0.834 & 0.894 \\
\cline{2-7}
 & BI-LSTM & 0.761 & 0.743 & 0.752 & 0.834 & 0.895 \\
\hline
\end{tabular}
\label{tab_all_results}
\end{center}
\end{table*}

\section{Turn Prediction Models}

To evaluate the effectiveness of the generated dataset, several classification models were implemented, from traditional machine learning algorithms to advanced deep learning methods. Specifically, we utilized Decision Tree (DT), Random Forest (RF), Logistic Regression (LR), and Bidirectional Long Short-Term Memory (BI-LSTM) architectures. In each turn transition, the final one-third of the text sequence was labeled as 1, while two distinct segments from the remaining portion were randomly selected and labeled as 0. The final dataset consists of 12,560 positive and 25,120 negative samples. To ensure comparability across model subsets, the training data was downsampled to match the smallest subset size. The resulting training set contained 1,306 positive and 2,696 negative samples. For text representation, the intfloat/multilingual-e5-large embedding model was applied, as it is highly effective at capturing the semantic and structural nuances of the Turkish language. Furthermore, the performance of each model was evaluated using a 5-fold cross-validation approach. The specific hyperparameters and configurations for each model are detailed below:

\begin{itemize}
    \item \textbf{DT:} $crit=Gini, split=Best, min\_split=2$    \item \textbf{LR:} $pen=L2, sol=lbfgs, iter=1000, C=1.0$
    \item \textbf{RF:} $est=100, crit=Gini, depth=None, boot=T$
    \item \textbf{LSTM:} $hid=384, opt=Adam, lr=0.001$ 
    \item \textbf{BI-LSTM:} $hid=384, opt=Adam, lr=0.001$
    \item \textbf{Ensemble (LR+RF):} $vote=Soft, weight=Equal$
\end{itemize}


\section{Conclusion}



Turn-taking remains a significant challenge in conversational AI. While many chatbots have been developed, predicting exactly when a user has finished speaking is still difficult. In this study, we created a natural Turkish dialogue dataset using various Qwen LLM models and temperature settings to address the lack of available resources for the Turkish language. We then evaluated this dataset using several machine learning and deep learning models. The experimental results, shown in Table \ref{tab_all_results}, demonstrate that the dataset provides a strong foundation for training turn-taking models. The BI-LSTM and Ensemble (LR+RF) models delivered the most balanced performance across all metrics. Specifically, the BI-LSTM model achieved the highest accuracy of 0.839, while the Ensemble model reached a peak AUC of 0.910. Notably, the slightly lower performance observed in the qwen3.5-397b-a17b subset suggests that the increased linguistic complexity and more natural conversational flow of advanced models make turn-taking points harder to predict for classifiers compared to simpler outputs.

Furthermore, the high performance of the LSTM-based architectures confirms that understanding the linguistic flow is essential for managing transitions in Turkish speech. While simpler models can identify potential turn-ending points, the advanced models provide the stability needed for natural interaction. These results demonstrate that the synthetic dataset might be useful for real-world dialogue predictions.

\vspace{12pt}


\begin{thebibliography}{00}
\bibitem{b7} E. Ekstedt and G. Skantze, ``TurnGPT: A Transformer-based Language Model for Predicting Turn-taking in Spoken Dialog,'' Findings of the Association for Computational Linguistics: EMNLP 2020, pp. 2981--2990, November 2020.

\bibitem{b9} E. Ekstedt and G. Skantze, ``Projection of Turn Completion in Incremental Spoken Dialogue Systems,'' Proceedings of the 22nd Annual Meeting of the Special Interest Group on Discourse and Dialogue, pp. 431--437, July 2021.

\bibitem{b10} M. Roddy, G. Skantze, and N. Harte, ``Multimodal Continuous Turn-Taking Prediction Using Multiscale RNNs,'' Proceedings of the 20th ACM International Conference on Multimodal Interaction, pp. 186--190, October 2018.

\bibitem{b11} G. Castillo-López, G. de Chalendar, and N. Semmar, ``A Survey of Recent Advances on Turn-taking Modeling in Spoken Dialogue Systems,'' Proceedings of the 15th International Workshop on Spoken Dialogue Systems Technology, pp. 254--271, May 2025.

\bibitem{b12} A. Cano, G. Perez, L. Merino, and R. Gomez, ``Towards Improving Turn-Taking in Social Robots Using Visual-Only Voice Activity Detection in Multimodal Dialogue Systems,'' Social Robotics + AI: 17th International Conference, ICSR+AI 2025, Proceedings, Part II, pp. 207--221, September 2025.

\bibitem{b13} Y. Lin, Y. Zheng, M. Zeng, and W. Shi, ``Predicting Turn-Taking and Backchannel in Human-Machine Conversations Using Linguistic, Acoustic, and Visual Signals,'' arXiv preprint arXiv:2505.12654, 2025.

\bibitem{b14} B. Schuppler, M. Hagmueller, J. A. Morales-Cordovilla, and H. Pessentheiner, ``GRASS: The Graz Corpus of Read and Spontaneous Speech,'' Proceedings of the Ninth International Conference on Language Resources and Evaluation (LREC'14), pp. 1465--1470, May 2014.


\end{thebibliography}
\end{document}